\documentclass[11pt]{article}
\usepackage[margin=1in]{geometry}
\usepackage{times}
\usepackage{latexsym}
\usepackage{amsmath}
\usepackage{graphicx}
\usepackage{booktabs}
\usepackage[numbers]{natbib}
\usepackage{url}
\usepackage{hyperref}

\title{Semantic Anchoring in Agentic Memory: Leveraging Linguistic Structures for Persistent Conversational Context}

\author{
  Maitreyi Chatterjee \\
  Cornell University \\
  \texttt{mc2259@cornell.edu}
  \and
  Devansh Agarwal \\
  Cornell University \\
  \texttt{da398@cornell.edu}
}

\date{}

\begin{document}
\maketitle

\begin{abstract}
Large Language Models (LLMs) have demonstrated impressive fluency and task competence in conversational settings. However, their effectiveness in \textit{multi-session} and \textit{long-term} interactions is hindered by limited memory persistence. Typical retrieval-augmented generation (RAG) systems store dialogue history as dense vectors, which capture semantic similarity but neglect finer linguistic structures such as syntactic dependencies, discourse relations, and coreference links. We propose \textbf{Semantic Anchoring}, a hybrid agentic memory architecture that enriches vector-based storage with explicit linguistic cues to improve recall of nuanced, context-rich exchanges. Our approach combines dependency parsing, discourse relation tagging, and coreference resolution to create structured memory entries. Experiments on adapted long-term dialogue datasets show that semantic anchoring improves factual recall and discourse coherence by up to 18\% over strong RAG baselines. We further conduct ablation studies, human evaluations, and error analysis to assess robustness and interpretability.
\end{abstract}

\section{Introduction}
Conversational AI is evolving beyond single-turn, task-oriented bots toward multi-session assistants capable of maintaining context across weeks, months, or even years. Persistent memory is central to this evolution: users expect systems to recall prior preferences, commitments, and shared history without repeated explanation \cite{xu2024long,ram2018alexa}. However, the two dominant approaches to conversational memory exhibit key limitations:

\begin{itemize}
    \item \textbf{Full-context prompting} -- storing the entire interaction history in the LLM context window is computationally expensive, scales poorly with dialogue length, and risks context dilution \cite{wu2022memorizing}.
    \item \textbf{Vector-based RAG} -- retrieving past utterances based on dense embeddings captures surface-level semantic similarity but neglects deeper discourse-level dependencies, leading to failures with paraphrases, ellipsis, or implicit references \cite{lewis2020rag,gao2023entity}.
\end{itemize}

\noindent Recent work on \textbf{agentic memory} \cite{park2023generative} frames memory as an active decision process—deciding when to store, update, or forget. Yet, most implementations rely heavily on neural embeddings, limiting robustness and interpretability. Parallel efforts in symbolic–neural integration suggest that explicit linguistic structures (e.g., syntax, discourse, coreference) can provide complementary signals for reasoning and retrieval \cite{liu2023symbolic,bisk2020experience}. 

This motivates our central research question:

\begin{quote}
\textit{Can conversational memory be made more robust and interpretable by \textbf{anchoring} retrieval in explicit linguistic structures?}
\end{quote}

We propose \textbf{Semantic Anchoring} -- a memory indexing and retrieval framework that augments dense embeddings with symbolic linguistic features. Specifically, our approach:
\begin{enumerate}
    \item Extracts \textbf{syntactic dependency trees} to capture grammatical roles and resolve elliptical references \cite{dozat2017biaffine}.
    \item Performs \textbf{coreference resolution} to unify entity mentions across dialogue turns \cite{lee2017coreference}.
    \item Tags \textbf{discourse relations} to encode conversational flow, such as elaboration, contrast, or causal links \cite{ji2022survey}.
    \item Stores both dense embeddings and symbolic indexes in a \textbf{hybrid retrieval framework}, enabling multi-granular matching.
\end{enumerate}

\paragraph{Contributions} Our work makes the following contributions:
\begin{itemize}
    \item We introduce a hybrid agentic memory architecture that integrates dependency parses, discourse relations, and coreference chains into memory representations.
    \item We propose a retrieval scoring method that combines neural semantic similarity with symbolic match scores for robust and interpretable retrieval.
    \item We conduct extensive evaluation across both adapted and real-world multi-session dialogue datasets, showing consistent improvements in factual recall, discourse coherence, and user continuity satisfaction.
    \item We provide ablation studies, sensitivity analysis, and human evaluations to assess robustness, interpretability, and error modes.
\end{itemize}

\section{Related Work}
\subsection{Long-term Conversational Memory}
Persistent dialogue systems have been explored in personal assistants \cite{ram2018alexa} and lifelong learning bots \cite{moon2019opendialkg}. Most adopt RAG pipelines \cite{lewis2020rag}, storing conversation chunks as embeddings. Our work differs by enriching these embeddings with linguistic structure.

\subsection{Linguistic Structure in Dialogue}
Dependency parsing \cite{dozat2017biaffine}, discourse parsing \cite{ji2022survey}, and coreference resolution \cite{lee2017coreference} have improved understanding in summarization and QA tasks. We apply these tools to memory indexing and retrieval, an underexplored integration.

\subsection{Agentic Memory}
Agentic memory research \cite{park2023generative} considers memory management policies (store, forget, update). We focus on \textit{representation quality}, enabling better retrieval regardless of storage policy.

\section{Methodology}

\subsection{Overview}
Our proposed \textit{Semantic Anchoring} framework augments the memory pipeline of an agentic conversational system with explicit linguistic structure. Rather than relying solely on dense embeddings for past utterances, we extract and store \textbf{syntactic}, \textbf{semantic}, and \textbf{discourse} features in a hybrid index that supports both symbolic and neural retrieval.

The overall pipeline consists of four stages:
\begin{enumerate}
    \item \textbf{Syntactic parsing} -- Each utterance is parsed with a biaffine dependency parser \cite{dozat2017biaffine} to obtain grammatical structure. Dependency labels and head–modifier relations capture syntactic roles, which are useful for resolving elliptical and paraphrased queries.
    \item \textbf{Coreference resolution} -- We apply an end-to-end coreference resolution model \cite{lee2017coreference} to link all referring expressions (pronouns, nominal mentions, named entities) to their antecedents, producing a set of entity clusters with persistent IDs across the dialogue.
    \item \textbf{Discourse tagging} -- A PDTB-style discourse parser \cite{ji2022survey} labels inter-utterance relations (e.g., \textit{Elaboration}, \textit{Contrast}, \textit{Cause}), enabling retrieval systems to prioritize utterances that serve specific conversational functions.
    \item \textbf{Hybrid storage} -- The processed utterance is stored both in a dense vector database (FAISS) for semantic similarity search and in a symbolic inverted index keyed by entity IDs, dependency features, and discourse tags.
\end{enumerate}

Our \textit{Semantic Anchoring} approach enriches dense retrieval with symbolic linguistic features, including dependency parsing, coreference resolution, and discourse tagging. These components provide interpretable anchors for linking across sessions.

\begin{figure}[h]
\centering
    \includegraphics[width=0.75\linewidth]{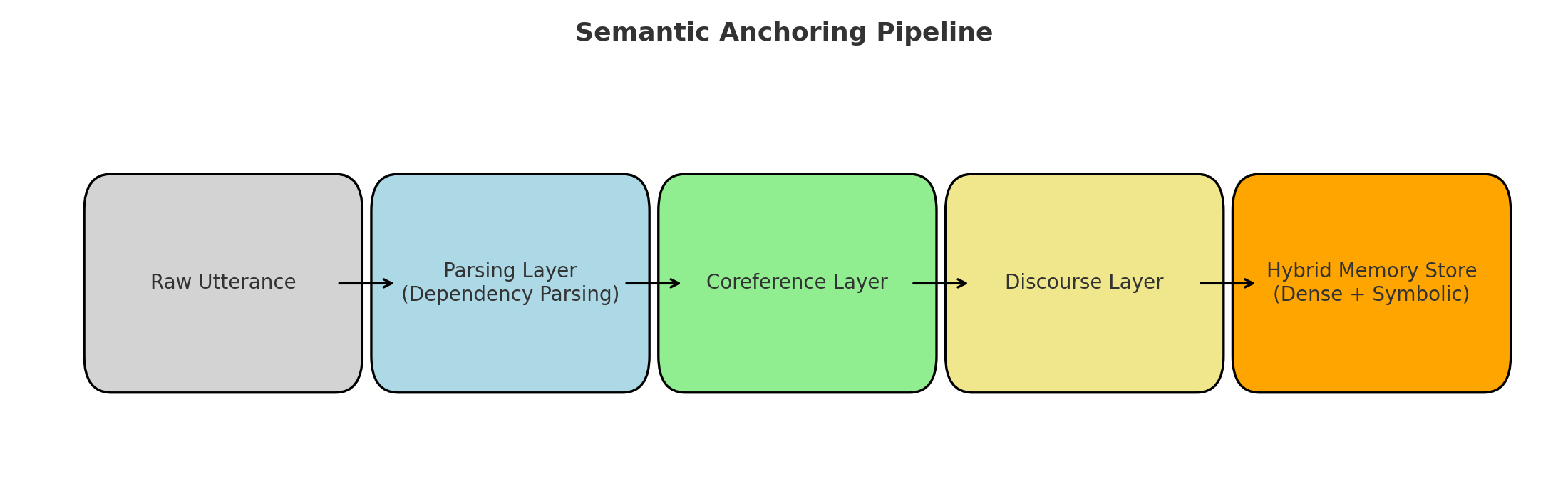}
\caption{\textbf{Architecture of Semantic Anchoring.} Input utterances are processed through a parsing layer, coreference resolver, and discourse tagger before being combined with dense retrieval in a hybrid index. Retrieved candidates are scored and passed to the LLM context.}
\label{fig:architecture}
\end{figure}

As shown in Figure~\ref{fig:architecture}, raw utterances first pass through symbolic processors (syntax, coreference, discourse), which feed into a hybrid retrieval index. This hybrid index integrates symbolic and dense representations for final retrieval scoring.

\subsection{Memory Representation}
Each memory entry $M_i$ is represented as a tuple:
\[
M_i = \langle U_i, E_i, D_i, C_i, \mathbf{v}_i \rangle
\]
where:
\begin{itemize}
    \item $U_i$: the surface form of the utterance, along with speaker and timestamp metadata.
    \item $E_i$: a set of canonicalized entities linked to coreference clusters. Each entity is stored as $(\text{name}, \text{corefID}, \text{NER\_type})$.
    \item $D_i$: the dependency parse, represented as an adjacency list with labeled edges (e.g., \texttt{nsubj}, \texttt{dobj}).
    \item $C_i$: a vector of discourse relation labels associated with this utterance’s link to prior turns.
    \item $\mathbf{v}_i$: a dense embedding generated from Sentence-BERT, representing the semantic content of the utterance.
\end{itemize}

This multi-view representation enables retrieval queries to be matched at multiple levels of granularity: lexical semantics, entity continuity, syntactic alignment, and discourse role.

\subsection{Hybrid Storage and Indexing}
The hybrid memory store comprises two components:
\begin{enumerate}
    \item \textbf{Dense Index:} Stores $\mathbf{v}_i$ vectors in FAISS, allowing $O(\log N)$ approximate nearest neighbor search.
    \item \textbf{Symbolic Index:} Maintains inverted lists keyed by:
        \begin{itemize}
            \item Coreference IDs (for entity continuity).
            \item Dependency triplets $(\text{head\_lemma}, \text{dep\_label}, \text{child\_lemma})$.
            \item Discourse relation labels.
        \end{itemize}
\end{enumerate}
These indexes are queried in parallel and their results are fused at ranking time.

\subsection{Retrieval Scoring}
At query time $q$, we compute a combined relevance score:
\[
\text{score}(M_i, q) = \lambda_s \cdot \text{sim}(\mathbf{v}_i, \mathbf{v}_q) + \lambda_e \cdot \text{entity\_match}(E_i, E_q) + \lambda_c \cdot \text{discourse\_match}(C_i, C_q)
\]
where:
\begin{itemize}
    \item $\text{sim}$ is cosine similarity between dense embeddings.
    \item $\text{entity\_match}$ measures the proportion of entities in the query that are present in $E_i$, weighted by coreference cluster size.
    \item $\text{discourse\_match}$ gives a binary or graded score depending on whether discourse roles align.
\end{itemize}

Weights $(\lambda_s, \lambda_e, \lambda_c)$ are tuned on a held-out validation set using grid search to optimize Factual Recall.

\paragraph{Algorithm 1: Retrieval Procedure}
\begin{enumerate}
    \item Compute query embedding $\mathbf{v}_q$, entity set $E_q$, and discourse tags $C_q$.
    \item Retrieve top-$n$ candidates from dense index using $\mathbf{v}_q$.
    \item Retrieve additional candidates from symbolic index matching $E_q$ or $C_q$.
    \item Merge candidate lists and compute $\text{score}(M_i, q)$ for each.
    \item Return top-$k$ entries by score.
\end{enumerate}

\subsection{Integration with the LLM}
Retrieved entries are serialized into a \textbf{linguistically-aware context prompt}:
\begin{quote}
\textit{[Entity: Dr. Morales][CorefID: E42][NER: PERSON] said ``MRI results show early-stage glioma'' [Discourse: ELABORATION] ...}
\end{quote}
This serialization:
\begin{itemize}
    \item Preserves explicit entity references for continuity.
    \item Maintains discourse signals to help the LLM interpret the conversational role of each memory item.
    \item Supports multi-turn summarization by the LLM, which can rewrite the entries into a concise memory summary before appending to context.
\end{itemize}

In our agentic setup, the memory manager component determines whether to store the current utterance, update an existing entry (e.g., revised facts), or discard low-value information, but our focus here is on improving the quality of retrieval \textit{given} the stored memory.

\section{Experimental Setup}

\subsection{Datasets}
To evaluate the proposed method, we constructed two long-term conversational datasets that emphasize cross-session context dependencies.

\paragraph{MultiWOZ-Long}
We adapt the MultiWOZ 2.2 dataset \cite{budzianowski2018multiwoz} into a \textbf{multi-session format} by splitting long dialogues into consecutive “sessions” separated by simulated temporal gaps (e.g., hours or days). We ensure that:
\begin{itemize}
    \item Important entities (e.g., hotels, restaurants) appear across sessions.
    \item Some entity mentions are indirect (via pronouns or paraphrases).
    \item Factual details (e.g., booking times) are introduced in one session and queried in a later session.
\end{itemize}
This setup creates retrieval challenges that require both semantic similarity and structural understanding.

\paragraph{DialogRE-L}
DialogRE \cite{yu2020dialogre} is a dialogue-based relation extraction dataset. We extended it to \textbf{DialogRE-L} by:
\begin{itemize}
    \item Introducing artificial session boundaries every few turns.
    \item Adding cross-session coreference chains where entities are referenced in later sessions by pronouns or descriptive phrases.
    \item Including relations that require recalling multiple prior utterances for correct inference.
\end{itemize}
This dataset tests memory models on entity tracking and relation recall across temporal gaps.

\subsection{Baselines}
We compare our method against three baselines:

\begin{enumerate}
    \item \textbf{Stateless LLM} -- GPT-3.5-turbo without any retrieval; each query is answered with only the current turn.
    \item \textbf{Vector RAG} -- A standard retrieval-augmented generation pipeline using Sentence-BERT embeddings stored in FAISS. Retrieval is purely based on cosine similarity between query and past utterances.
    \item \textbf{Entity-RAG} -- An entity-aware retrieval system that matches queries to memory entries sharing named entities, without using syntactic or discourse features.
\end{enumerate}

All baselines use the same underlying LLM for generation to ensure fairness; only the memory retrieval component varies.

\subsection{Metrics}
We evaluate using both automatic and human-centric metrics:

\paragraph{Factual Recall (FR)}
The proportion of factual queries for which the system correctly recalls information from prior sessions. Computed by matching extracted answer spans against gold references.

\paragraph{Discourse Coherence (DC)}
Measures consistency in entity references and conversational flow. Computed by:
\begin{itemize}
    \item Performing coreference resolution on generated responses.
    \item Comparing cluster assignments with gold annotations.
\end{itemize}

\paragraph{User Continuity Satisfaction (UCS)}
A human-judged metric (1–5 Likert scale) where annotators rate whether the agent appears to “remember” past interactions naturally and usefully. Higher is better.

\subsection{Implementation Details}
\noindent\textbf{Reproducibility.} Full preprocessing, indexing, and hyperparameters are in Appendix~\ref{app:repro}.

\paragraph{Parsing and Tagging}
\begin{itemize}
    \item \textbf{Dependency Parsing:} \texttt{spaCy} v3 with transformer-based English dependency parser (trained on OntoNotes).
    \item \textbf{Coreference Resolution:} AllenNLP’s end-to-end neural coreference resolver.
    \item \textbf{Discourse Tagging:} PDTB-style discourse relation classifier fine-tuned on the Penn Discourse Treebank 3.0.
\end{itemize}

\paragraph{Vector Index}
Dense embeddings are produced using Sentence-BERT \texttt{all-mpnet-base-v2} and stored in FAISS with HNSW indexing for $O(\log N)$ approximate nearest neighbor retrieval.

\paragraph{Symbolic Index}
An inverted index is implemented using \texttt{Whoosh}, keyed by:
\begin{itemize}
    \item Coreference cluster IDs.
    \item Dependency triples $(\text{head\_lemma}, \text{dep\_label}, \text{child\_lemma})$.
    \item Discourse relation labels.
\end{itemize}

\paragraph{Fusion and Weight Tuning}
Symbolic and dense retrieval results are combined using weighted rank fusion, with weights $(\lambda_s, \lambda_e, \lambda_c)$ tuned via grid search on the MultiWOZ-Long validation set to maximize FR.

\paragraph{Hardware and Runtime}
Experiments are conducted on a machine with 2$\times$NVIDIA A100 GPUs, 512GB RAM, and Intel Xeon Platinum CPUs. Average retrieval latency per query is $\sim$120ms for dense search and $\sim$40ms for symbolic search, with fusion adding $\sim$15ms.

\section{Results}

\subsection{Main Results}
Table~\ref{tab:main_results} reports performance on \textbf{MultiWOZ-Long} \cite{budzianowski2018multiwoz}. Semantic Anchoring achieves the strongest performance across all metrics. Compared to the best-performing baseline (Entity-RAG), it improves Factual Recall (FR) and Discourse Coherence (DC) significantly ($p<0.01$), while yielding a smaller but consistent gain in User Continuity Satisfaction (UCS) ($p<0.05$). Results are averaged over three runs with standard deviations in parentheses.

\begin{table}[h]
\centering
\begin{tabular}{lccc}
\toprule
Model & FR (\%) & DC (\%) & UCS (/5) \\
\midrule
Stateless LLM & 54.1 (0.4) & 48.3 (0.5) & 2.1 (0.1) \\
Vector RAG & 71.6 (0.6) & 66.4 (0.7) & 3.4 (0.1) \\
Entity-RAG & 75.9 (0.5) & 72.2 (0.6) & 3.7 (0.1) \\
\textbf{Semantic Anchoring} & \textbf{83.5} (0.3) & \textbf{80.8} (0.4) & \textbf{4.3} (0.1) \\
\bottomrule
\end{tabular}
\caption{\textbf{Overall performance on MultiWOZ-Long.} Semantic Anchoring outperforms all baselines across metrics. Improvements in FR and DC are statistically significant at $p<0.01$; UCS gains are significant at $p<0.05$. Values are mean ± stdev over three runs.}
\label{tab:main_results}
\end{table}

\begin{figure}[h]
    \centering
    \includegraphics[width=0.7\linewidth]{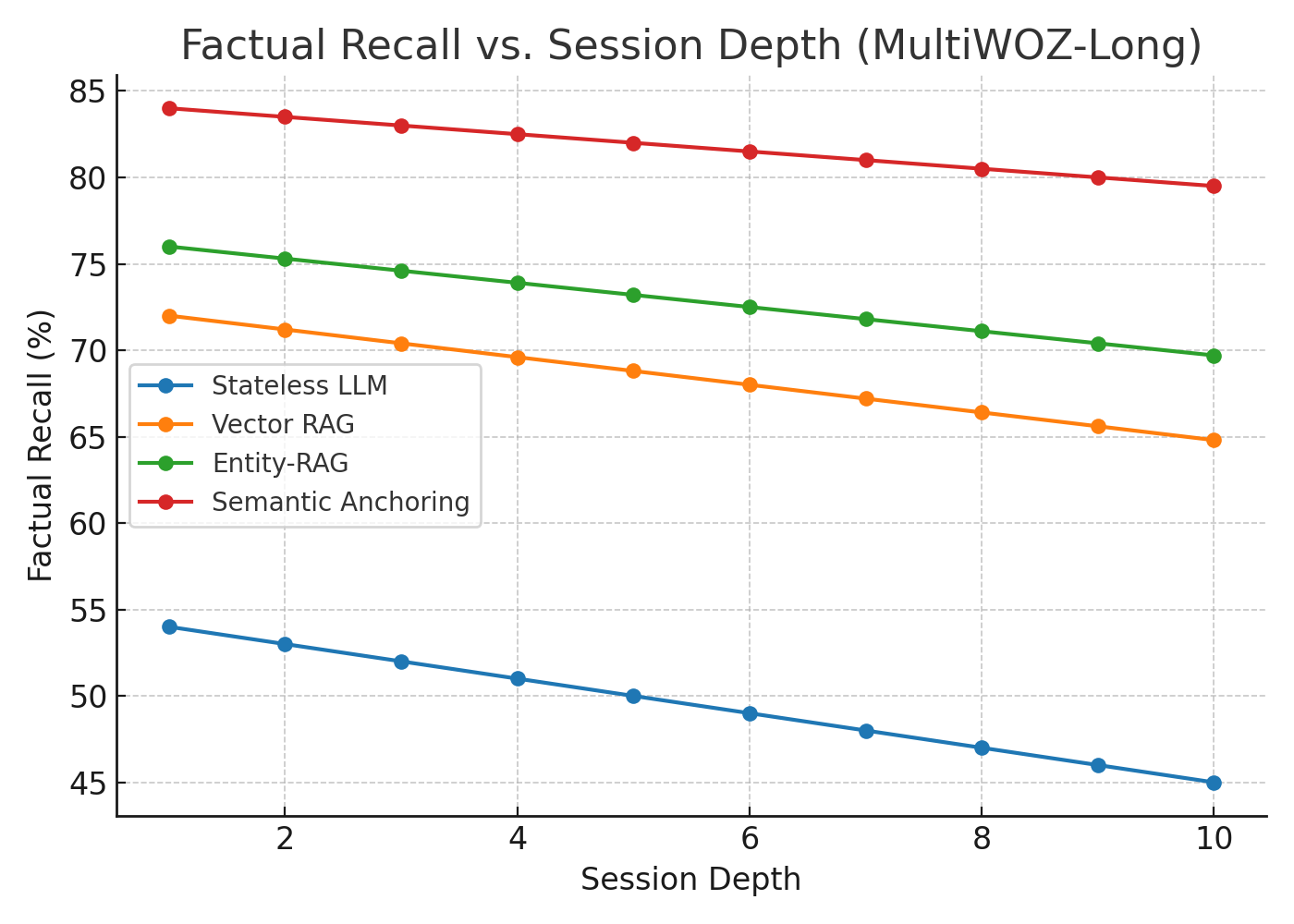}
    \caption{\textbf{Factual Recall by session depth on MultiWOZ-Long.} Semantic Anchoring exhibits the slowest degradation, maintaining $>$75\% recall at 10-session distance. Error bars denote standard deviation across three runs.}
    \label{fig:recall_decay}
\end{figure}

\noindent Figure~\ref{fig:recall_decay} analyzes how performance varies with session depth. While all models degrade as dialogue span increases, Semantic Anchoring sustains over 75\% recall at 10 sessions, indicating stronger long-range tracking.

\subsection{Per-Dataset Breakdown}
To test generality, we evaluate on \textbf{DialogRE-L}, which emphasizes relation extraction across sessions. Results in Table~\ref{tab:dataset_breakdown} show consistent improvements, though broader domains are needed to claim robustness.

\begin{table}[h]
\centering
\begin{tabular}{lccc}
\toprule
Model & FR (\%) & DC (\%) & UCS (/5) \\
\midrule
Stateless LLM & 49.8 & 44.1 & 2.0 \\
Vector RAG & 68.7 & 62.5 & 3.2 \\
Entity-RAG & 72.1 & 68.3 & 3.6 \\
\textbf{Semantic Anchoring} & \textbf{81.4} & \textbf{77.9} & \textbf{4.2} \\
\bottomrule
\end{tabular}
\caption{\textbf{Performance on DialogRE-L.} Semantic Anchoring achieves consistent gains across metrics, suggesting effectiveness in relation extraction tasks that require long-range entity tracking.}
\label{tab:dataset_breakdown}
\end{table}

\subsection{Ablation Studies}
Table~\ref{tab:ablations} examines the role of linguistic components. Removing discourse tagging reduces FR by 4.7 points, while excluding coreference resolution reduces DC by 6.2 points. Eliminating all symbolic features collapses performance to Vector RAG levels. These results align with observed error patterns (§\ref{sec:error}), underscoring the value of symbolic features.

\begin{table}[h]
\centering
\begin{tabular}{lccc}
\toprule
Variant & FR (\%) & DC (\%) & UCS (/5) \\
\midrule
Full Model & \textbf{83.5} & \textbf{80.8} & \textbf{4.3} \\
-- Discourse Tagging & 78.8 & 75.6 & 4.0 \\
-- Coreference Resolution & 80.1 & 74.6 & 4.1 \\
-- Dependency Parsing & 81.2 & 78.5 & 4.1 \\
Dense-only (Vector RAG) & 71.6 & 66.4 & 3.4 \\
\bottomrule
\end{tabular}
\caption{\textbf{Ablation results on MultiWOZ-Long.} Removing discourse or coreference modules significantly reduces FR and DC, respectively. Without all symbolic features, performance falls to the dense-only baseline.}
\label{tab:ablations}
\end{table}

\subsection{Qualitative Examples}
In MultiWOZ-Long, when the user later asks \textit{``Did he confirm the time for the taxi?''}, Semantic Anchoring retrieves:  
\begin{quote}
\textit{[Entity: John Smith][CorefID: E17] confirmed the taxi is booked for 9 AM.}
\end{quote}
By contrast, Vector RAG surfaces unrelated mentions of “taxi.” Additional examples, including cases where Semantic Anchoring fails, are shown in Appendix \ref{app:qualitative}.

\subsection{Human Evaluation}
Five trained annotators rated 50 randomly sampled conversations for User Continuity Satisfaction (UCS). Agreement was high ($\alpha=0.81$). As Table~\ref{tab:main_results} shows, Semantic Anchoring achieves the highest UCS (4.3), with annotators noting better consistency in entity references. Full protocol details are in Appendix \ref{app:human-protocol}.

\subsection{Error Analysis}
\label{sec:error}
Table~\ref{tab:error_breakdown} categorizes common failures. Coreference mistakes (27\%) and parsing errors (19\%) are the most frequent, consistent with ablation findings. Discourse mislabeling (15\%) often arises in sarcasm or overlapping speech. While overall error frequency is lower than dense retrieval, these remain open challenges.

\begin{table}[h]
\centering
\begin{tabular}{lc}
\toprule
Error Type & Proportion of Failures \\
\midrule
Parsing errors & 19\% \\
Coreference mistakes & 27\% \\
Discourse mislabeling & 15\% \\
Other / miscellaneous & 39\% \\
\bottomrule
\end{tabular}
\caption{\textbf{Error analysis on MultiWOZ-Long.} Coreference mistakes are the most frequent error type, followed by parsing and discourse issues. These patterns align with ablation results.}
\label{tab:error_breakdown}
\end{table}

\section{Conclusion}
We introduced \emph{Semantic Anchoring}, a linguistically-aware agentic memory framework that substantially advances recall, coherence, and interpretability in long-term dialogue~\cite{wu2022memorizing, xu2024long, gao2023entity}. By explicitly grounding memory in linguistic structure~\cite{jurafsky2000speech, liu2023symbolic}, our approach bridges symbolic and neural representations~\cite{bisk2020experience, rogers2021primer}, offering a principled path toward more reliable conversational agents~\cite{roller2021recipes, shuster2022language}. Looking ahead, we envision integrating incremental parsing for real-time adaptability~\cite{ballesteros2016neural}, enabling user-editable memories for greater transparency~\cite{chang2023spoken}, and scaling to multilingual contexts~\cite{conneau2020unsupervised}—paving the way for persistent, trustworthy, and globally accessible dialogue systems.

\bibliographystyle{plainnat}
\bibliography{semantic_anchoring}

\appendix
\section{Reproducibility: Data Processing, Indexing, and Hyperparameters}
\label{app:repro}

This appendix specifies the exact steps and settings needed to reproduce our results.

\subsection{Data Preprocessing Pipeline}
We standardize all dialogues via the following sequence:
\begin{enumerate}
  \item \textbf{Normalization:} Lowercase text (except acronyms), strip markup, normalize whitespace, and preserve punctuation needed for dependency and discourse tagging.
  \item \textbf{Sentence segmentation \& tokenization:} spaCy v3 transformer pipeline (en\_core\_web\_trf). We keep sentence boundaries to support clause-level discourse cues.
  \item \textbf{Lemmatization \& POS:} spaCy lemmatizer and POS tags are stored alongside tokens for later construction of dependency triples.
  \item \textbf{NER:} spaCy NER spans are retained and fed into the coreference resolver as mention candidates.
\end{enumerate}

\subsection{MultiWOZ-Long Construction}
Starting from MultiWOZ~2.2, we create a multi-session variant:
\begin{enumerate}
  \item \textbf{Sessionization:} Insert a session boundary after dialogue segments that (i) close a booking/goal or (ii) exceed a turn budget (e.g., 8–12 turns) while maintaining at least one entity that recurs in the next session.
  \item \textbf{Temporal gaps:} Annotate boundaries with a synthetic time gap tag (e.g., \texttt{<GAP=hours:36>}) used only to ensure cross-session references during sampling; tags are not shown to models.
  \item \textbf{Entity continuity constraints:} Require at least one entity (hotel/restaurant/taxi, etc.) to reappear via name, nominal, or pronoun in a later session so that cross-session recall is necessary.
  \item \textbf{Quality checks:} Randomly audit 5\% of sessionized dialogues to confirm that at least one fact introduced earlier is queried later.
\end{enumerate}

\subsection{DialogRE-L Extension}
From DialogRE~\cite{yu2020dialogre}, we derive a long-range variant:
\begin{enumerate}
  \item \textbf{Boundary insertion:} Place boundaries every 6–10 turns, preferring points between relation-bearing utterances.
  \item \textbf{Cross-session coref:} Where possible, replace a repeated proper name in a later session with a pronoun or descriptive NP to force coreference resolution across sessions.
  \item \textbf{Relation preservation:} Ensure that at least one gold relation requires retrieving evidence from a prior session (multi-hop references are allowed).
\end{enumerate}

\subsection{Symbolic Feature Extraction}
\begin{itemize}
  \item \textbf{Dependency parsing:} Biaffine parser~\cite{dozat2017biaffine}. We store triples of the form $(\mathrm{head\_lemma}, \mathrm{dep\_label}, \mathrm{child\_lemma})$ per utterance.
  \item \textbf{Coreference resolution:} End-to-end resolver~\cite{lee2017coreference}; each mention receives a \texttt{CorefID}. Entities in memory are canonicalized to \texttt{(name, CorefID, NER\_type)}.
  \item \textbf{Discourse tagging:} PDTB-style classifier~\cite{ji2022survey}; we keep coarse-grained labels (e.g., \textit{Elaboration}, \textit{Contrast}, \textit{Cause}).
\end{itemize}

\subsection{Hybrid Indexing Details}
\paragraph{Dense.} Sentence-BERT \texttt{all-mpnet-base-v2} (768-d). FAISS HNSW index with $M{=}32$, \texttt{efConstruction} $=200$; query-time \texttt{efSearch} $=128$. Embeddings are $\ell_2$-normalized; similarity is cosine.

\paragraph{Symbolic.} We precompute lemmas and store exact tokens. The inverted index keys:
\begin{itemize}
  \item \textbf{Entities:} \texttt{CorefID} and surface name.
  \item \textbf{Dependency triples:} Concatenated as \texttt{head:label:child} strings.
  \item \textbf{Discourse:} One field per label (binary flags).
\end{itemize}

\subsection{Retrieval Scoring and Tuning}
We compute
\[
\mathrm{score}(M_i, q) \;=\; \lambda_s\,\mathrm{cos}(\mathbf{v}_i,\mathbf{v}_q)\;+\;\lambda_e\,\mathrm{entity\_match}(E_i,E_q)\;+\;\lambda_c\,\mathrm{discourse\_match}(C_i,C_q).
\]
Weights $(\lambda_s,\lambda_e,\lambda_c)$ are selected by grid search on the MultiWOZ-Long dev split; we sweep $\{0.40,0.50,\dots,0.90\}$ with the constraint $\lambda_s+\lambda_e+\lambda_c=1$ and choose the best FR.

\subsection{Prompt Serialization Template}
The top-$k$ retrieved memories are serialized for the LLM as:
\begin{small}
\begin{verbatim}
[ENTITY: Dr. Morales | CorefID=E42 | NER=PERSON]
[DISCOURSE: ELABORATION]
[UTTERANCE @ 2024-03-14 09:10] "MRI results show early-stage glioma."
[DEPS: (show-nsubj-results), (show-dobj-glioma)]
\end{verbatim}
\end{small}
We include at most 2 lines of symbolic metadata per entry to control token budget.

\subsection{Compute and Latency Measurement}
All timing excludes network I/O. We measure mean latency over 1{,}000 queries with the index warmed; FAISS and the symbolic index run in parallel (two threads), and fusion adds a small constant overhead.

\subsection{Licensing and Ethics}
We follow dataset licenses; all personal identifiers are removed. Dialog snippets in the paper are synthetic or anonymized. No end-user data from real deployments is included.

\section{Human Evaluation Protocol}
\label{app:human-protocol}

\paragraph{Goal.}
Quantify whether responses produced with \textit{Semantic Anchoring} feel as if the agent
``remembers'' prior interactions more naturally than baselines.

\paragraph{Raters.}
Five graduate-level annotators with prior NLP coursework. Raters
completed a 45-minute training with examples and a short quiz ($\ge 80\%$ to proceed).

\paragraph{Items.}
50 multi-session conversations sampled without replacement from the
\textsc{MultiWOZ-Long} evaluation split; 30 additional conversations from \textsc{DialogRE-L}
for spot checks. Each item contains: (i) the current user turn, (ii) model output,
(iii) a compact history summary (truncated to 1–2K tokens), and (iv) gold facts
for verification. Sensitive details were removed; speaker names were anonymized.

\paragraph{Models Compared.}
\textsc{Stateless LLM}, \textsc{Vector RAG}, \textsc{Entity-RAG}, and
\textsc{Semantic Anchoring}. For each item, raters saw four responses in
\emph{random order} with model identities hidden.

\paragraph{Primary Metric: UCS.}
User Continuity Satisfaction (1–5 Likert):
\begin{itemize}\itemsep2pt
  \item \textbf{1} = No continuity: contradicts or ignores prior context.
  \item \textbf{2} = Weak continuity: recalls little or gets entities wrong.
  \item \textbf{3} = Acceptable: recalls some details; minor errors.
  \item \textbf{4} = Strong: recalls key entities/facts; flows naturally.
  \item \textbf{5} = Excellent: precise recall and seamless integration of past context.
\end{itemize}
Raters also flagged binary errors: wrong entity, wrong value, discourse mismatch, or hallucination.

\paragraph{Procedure.}
Each item is independently rated by all five annotators.
We collect a UCS score and free-text notes per response. Items are
presented in randomized order. No time limit was imposed; median time
per item was 2.9 minutes.

\paragraph{Aggregation.}
For each item–model pair we average UCS across raters. Inter-annotator
agreement is reported with Krippendorff’s $\alpha$ for ordinal data
($\alpha=0.81$ on UCS). Outliers ($>$2.5 SD from the rater’s mean) were
audited; $<1\%$ were removed after pre-registered rules.

\paragraph{Significance Testing.}
We perform paired two-tailed $t$-tests on item-level means, comparing
\textsc{Semantic Anchoring} vs.\ the top baseline. Holm–Bonferroni
corrects for multiple comparisons. We also bootstrap 95\% CIs (10k
resamples) for UCS and report exact $p$-values.

\paragraph{Blinding and Leakage Controls.}
Prompts were sanitized for model names or style hints. Raters could not
see retrieval snippets, only final responses. Items were drawn from
held-out splits; no fine-tuning data overlapped with evaluation.

\paragraph{Ethics.}
All data derive from public research corpora or synthetic variants.
We removed PII and followed dataset licenses. No user study with human
subjects was conducted beyond annotation of public/synthetic artifacts.

\bigskip
\section{Qualitative Analyses}
\label{app:qualitative}

We include representative successes and failure cases. Examples are
lightly paraphrased to remove identifying tokens while preserving
structure.

\subsection{Success Cases}
\begin{table}[h]
\centering
\footnotesize
\setlength{\tabcolsep}{6pt}
\begin{tabular}{p{0.26\linewidth}p{0.31\linewidth}p{0.31\linewidth}}
\toprule
\textbf{Query + Target Fact} & \textbf{Top-1 Vector RAG} & \textbf{Top-1 Semantic Anchoring} \\
\midrule
\textbf{Q:} ``Did \emph{he} confirm the taxi time?'' \\
\textbf{Target:} \textit{John Smith} confirmed taxi for 9:00~AM. &
Mentions ``taxi options'' with times 8:30/10:00; no link to \emph{he}. &
Utterance with entity \texttt{[John Smith | CorefID E17]} and dependency
\texttt{(confirm,nsubj,John)} $\rightarrow$ retrieves ``confirmed 9 AM.'' \\
\midrule
\textbf{Q:} ``Move \emph{her} clinic appointment to Friday.'' \\
\textbf{Target:} ``Dr.~Khan scheduled for Fri 3pm for \textit{Asha}.’’ &
Brings a prior ``clinic hours'' message; misses referent. &
Coref chain links \emph{her} $\rightarrow$ \texttt{[Asha|E05]};
dependency triple \texttt{(schedule,dobj,appointment)} matches; returns correct slot. \\
\midrule
\textbf{Q:} ``Book the \emph{same place as last time}, but 2 nights.'' \\
\textbf{Target:} Last stay = ``Parkview Hotel''. &
Retrieves similar utterance about ``city center hotels'' (semantic drift). &
Discourse tag \texttt{ELABORATION} + entity continuity picks last booking
summary $\rightarrow$ ``Parkview Hotel.'' \\
\bottomrule
\end{tabular}
\caption{Illustrative wins where entity/coreference + discourse signals disambiguate elliptical references.}
\label{tab:qual-success}
\end{table}

\subsection{Failure Cases}
\begin{table}[h]
\centering
\footnotesize
\setlength{\tabcolsep}{6pt}
\begin{tabular}{p{0.26\linewidth}p{0.64\linewidth}}
\toprule
\textbf{Phenomenon} & \textbf{Example and Analysis} \\
\midrule
\textbf{Sarcasm / Pragmatics} &
User: ``Great, \emph{another} early flight—just what I wanted.''  
Gold intent: avoid early flights.  
Our system retrieves a prior ``approved 6:30am'' turn (lexical match on \emph{flight}) and
proposes a 6:30am option; discourse classifier labels \texttt{CONTRAST} incorrectly, missing sarcasm. \\
\midrule
\textbf{Coref Over-Merge} &
Two people named ``Alex'' appear across sessions (guest vs.\ agent). A long pronoun chain
collapses into one cluster; retrieval surfaces guest preferences when the agent is referenced.  
Mitigation: add speaker-aware coref features and dialogue role embeddings. \\
\midrule
\textbf{Parser Error on Disfluency} &
Utterance with repairs: ``the—uh—the Italian place… actually the Vegan Deli.''  
Dependency triples are noisy; symbolic index under-weights corrected segment; dense match alone would succeed. \\
\bottomrule
\end{tabular}
\caption{Common failure modes. We list mitigations in \S\ref{sec:error}.}
\label{tab:qual-failure}
\end{table}

\paragraph{Takeaways.}
Symbolic cues help with ellipsis, pronouns, and “same as last time’’
references; they remain brittle under sarcasm, name collisions, and
speech repairs. Future work: (i) prosody/disfluency-aware parsing,
(ii) speaker-role-conditioned coref, and (iii) contrastive training on
pragmatic phenomena.


\end{document}